\documentclass[11pt]{article}

\usepackage[final]{acl}

\usepackage[utf8]{inputenc}
\usepackage[T1]{fontenc}

\usepackage{times}
\usepackage{amsmath}
\usepackage{latexsym}
\usepackage{microtype}
\usepackage{inconsolata}
\usepackage{graphicx}
\usepackage{booktabs}
\usepackage{pgfplots}
\pgfplotsset{compat=1.17}

\title{BUSTED at AraGenEval Shared Task: A Comparative Study of Transformer-Based Models for Arabic AI-Generated Text Detection}

\author{
Ali Zain \\
{\tt vin.alizain@gmail.com} 
\And
Sareem Farooqui \\
{\tt sareemfarooqui10@gmail.com} 
\AND
Muhammad Rafi \\
{\tt muhammad.rafi@nu.edu.pk} \\ \\
National University of Computer and Emerging Sciences, FAST \\
Karachi, Pakistan
}

\begin{document}
\maketitle
\begin{abstract}
    This paper details our submission to the AraGenEval Shared Task on Arabic AI-generated text detection, where our team, BUSTED, secured 5th place. We investigated the effectiveness of three pre-trained transformer models: AraELECTRA, CAMeLBERT, and XLM-RoBERTa. Our approach involved fine-tuning each model on the provided dataset for a binary classification task. Our findings revealed a surprising result: the multilingual XLM-RoBERTa model achieved the highest performance with an F1 score of 0.7701, outperforming the specialized Arabic models. This work underscores the complexities of AI-generated text detection and highlights the strong generalization capabilities of multilingual models.
\end{abstract}

\section{Introduction}
The increasing sophistication of large language models (LLMs) has blurred the line between human and machine-authored text. This reality poses significant societal risks, from accelerating the spread of misinformation to undermining academic integrity. In response, the development of reliable detectors for AI-generated text has become a pressing research priority. The AraGenEval Shared Task \cite{abudalfa2025arageneval} provides a crucial benchmark for this challenge in the Arabic language, a domain where such tools are still developing.

Our approach was to systematically evaluate the performance of different transformer architectures. We fine-tuned each model to perform binary classification, adapting their general linguistic knowledge to the specific task of distinguishing human from machine authorship. We specifically investigated:
\begin{enumerate}
    \item \textbf{AraELECTRA} \cite{antoun2021araelectra}, a specialized Arabic model.
    \item \textbf{CAMeLBERT} \cite{inoue2021interplay}, a widely-used Arabic BERT model.
    \item \textbf{XLM-RoBERTa} \cite{conneau2019unsupervised}, a large multilingual model.
\end{enumerate}

This paper's contributions are threefold. First, we provide a direct comparison of monolingual versus multilingual models for Arabic text detection. Second, we demonstrate that a multilingual model can achieve superior performance, a counter-intuitive but important finding. Finally, we analyze how certain preprocessing choices, such as aggressive text normalization, can inadvertently harm model performance by erasing subtle stylistic cues. Our best-performing model secured a 5th place finish in the shared task.

\section{Related Work}
Early efforts in authorship attribution and machine-text detection relied on statistical stylometry, using features like n-gram frequencies, readability scores, and syntactic structures to train classifiers. While effective for simpler models, these methods are less robust against the fluency of modern LLMs.

The current research landscape is dominated by neural network approaches. Fine-tuning pre-trained transformers like BERT \cite{devlin2018bert} has emerged as a powerful and accessible baseline. Other lines of inquiry focus on detecting statistical artifacts unique to the generative process of LLMs or embedding a "watermark" into the text during generation. Recent work has also shown the efficacy of comparing multiple strategies, from fine-tuned transformers like RoBERTa \cite{busted2025multi}. Our work aligns with the fine-tuning paradigm and is inspired by comprehensive comparative studies like that of \cite{alshboul2024comprehensive}, applying a similar methodology to the specific and under-resourced domain of Arabic AI-text detection.

\section{Background}
\subsection{Task Setup}
The AraGenEval shared task is a binary text classification problem. The goal is to classify a given Arabic text snippet as either `human-written` or `machine-generated`.
\begin{itemize}
    \item \textbf{Input}: A string of Arabic text.
    \item \textbf{Output}: A binary label (`human` or `machine`).
\end{itemize}

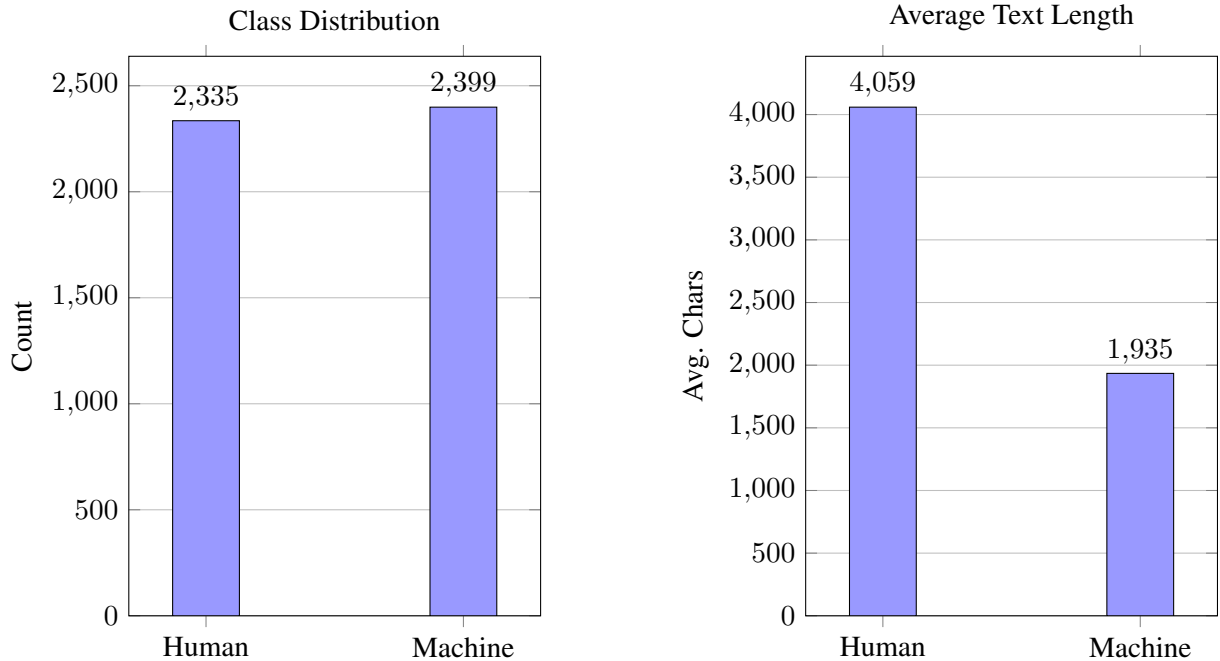
\begin{figure*}[t]
    \centering
    \begin{tikzpicture}
        \begin{axis}[
            ybar,
            bar width=25pt,
            width=7cm, height=9cm,
            enlarge x limits=0.3,
            ylabel={Count},
            symbolic x coords={Human, Machine},
            xtick=data,
            nodes near coords,
            nodes near coords align={vertical},
            ymin=0,
            title={Class Distribution},
            ymajorgrids=true,
        ]
        \addplot[fill=blue!40] coordinates {(Human,2335) (Machine,2399)};
        \end{axis}
    \end{tikzpicture}
    \hfill
    \begin{tikzpicture}
        \begin{axis}[
            ybar,
            bar width=25pt,
            width=7cm, height=9cm,
            enlarge x limits=0.3,
            ylabel={Avg. Chars},
            symbolic x coords={Human, Machine},
            xtick=data,
            nodes near coords,
            nodes near coords style={
                /pgf/number format/precision=0,
                /pgf/number format/fixed
            },
            ymin=0,
            title={Average Text Length},
            ymajorgrids=true,
        ]
        \addplot[fill=blue!40] coordinates {(Human,4059.13) (Machine,1934.53)};
        \end{axis}
    \end{tikzpicture}
    \caption{Statistics of the AraGenEval training dataset. 
    The classes are well-balanced, but human-written texts are 
    more than twice as long as machine-generated ones.}
    \label{fig:dataset_stats}
\end{figure*}

\subsection{Dataset Analysis}
The task utilized the AraGenEval dataset, which, after cleaning, contains 4,734 training samples. The class distribution is nearly balanced, with 2,399 samples (50.68\%) labeled as 'machine' and 2,335 (49.32\%) as 'human'. Our initial analysis revealed several key distinguishing features within the training data:

\paragraph{Text Length:} A significant discriminator is text length. Human-written texts are substantially longer on average (4059.13 characters) compared to machine-generated texts (1934.53 characters). This suggests that document length alone could be a strong, albeit potentially brittle, feature.

\paragraph{Lexical and N-gram Differences:} We observed distinct topical and stylistic patterns.
\begin{itemize}
    \item \textbf{Human-written texts} frequently contain words like ``Gaza'', ``the war'', and ``Israel'', and n-grams such as ``the United States'', pointing to a focus on specific current geopolitical events.
    \item \textbf{Machine-generated texts} use more general and formal vocabulary, such as ``can be'', ``in a way'', and n-grams like ``the international community'' and ``human rights'', suggesting a more analytical or descriptive style.
\end{itemize}
These lexical and phraseological differences highlight the distinct registers and topics between the two classes, which are crucial for classification.

\subsection{Related Work}
Our work is built on the transformer architecture \cite{vaswani2017attention}. Our comparative approach, which evaluates multiple deep learning models for an Arabic text classification task, is inspired by comprehensive surveys in the field, such as the one conducted by \cite{alshboul2024comprehensive}. We specifically leverage pre-trained models including BERT \cite{devlin2018bert}, ELECTRA \cite{clark2020electra}, and XLM-RoBERTa \cite{conneau2019unsupervised}. Our chosen models, CAMeLBERT \cite{inoue2021interplay} and AraELECTRA \cite{antoun2021araelectra}, are state-of-the-art for the Arabic language, while XLM-RoBERTa is a robust multilingual baseline.

\section{System Overview}
\begin{figure}[t]
    \centering
    \includegraphics[width=\columnwidth,height=6cm,keepaspectratio]{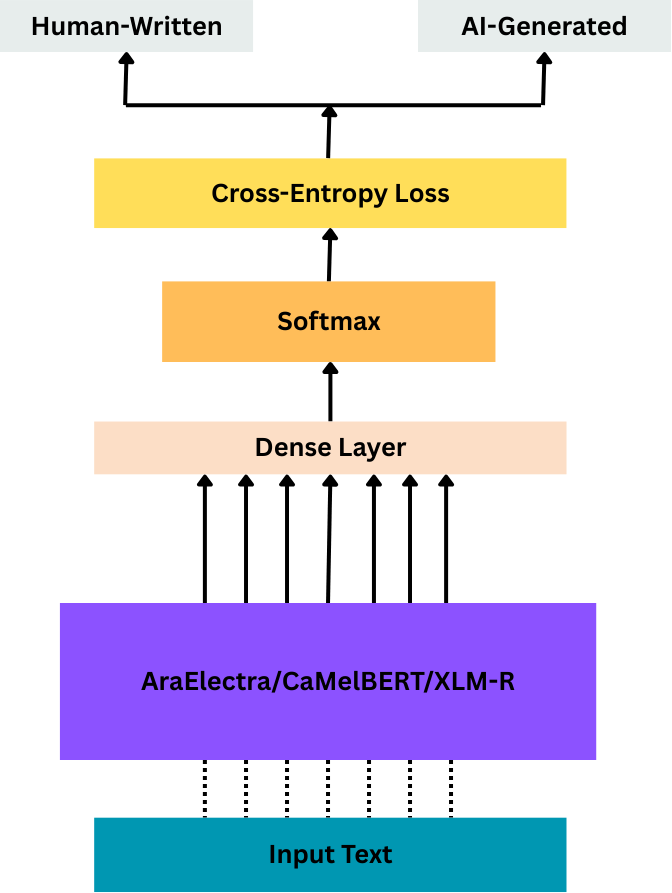}
    \caption{Overview of our comparative system. Input text is processed in parallel by three separate fine-tuned models. AraELECTRA's pipeline includes an additional text normalization step.}
    \label{fig:system_diagram}
\end{figure}

We implemented three systems based on different pre-trained models. Our overall workflow is illustrated in Figure \ref{fig:system_diagram}.

\subsection{System 1: AraELECTRA}
This system uses `aubmindlab/araelectra-base-discriminator`. A key component was an aggressive Arabic text normalization preprocessing step applied before tokenization. This function normalized various Arabic characters (e.g., alef variants to standard alef, and ta marbuta to ha) and stripped all Arabic diacritics and non-alphanumeric characters.

\subsection{System 2: CAMeLBERT}
This system is based on `CAMeL-Lab/bert-base-arabic-camelbert-mix`. In contrast to the AraELECTRA system, we did not apply any specific text normalization, relying entirely on the model's pre-trained tokenizer.

\subsection{System 3: XLM-RoBERTa}
Our third and best-performing system utilizes the multilingual `xlm-roberta-base` model. Similar to the CAMeLBERT setup, no language-specific normalization was performed.
\begin{table*}[ht]
\centering
\small
\begin{tabular}{@{}lcccccc@{}}
\toprule
\textbf{Model} & \textbf{F1-Score} & \textbf{Accuracy} & \textbf{Precision} & \textbf{Recall} & \textbf{Specificity} & \textbf{Balanced Acc.} \\ \midrule
XLM-RoBERTa & \textbf{0.7701} & \textbf{0.760} & \textbf{0.7390} & \textbf{0.804} & \textbf{0.716} & \textbf{0.760} \\
CAMeLBERT   & 0.7290 & 0.710 & 0.6842 & 0.780 & 0.640 & 0.710 \\
AraELECTRA  & 0.6180 & 0.550 & 0.5369 & 0.728 & 0.372 & 0.550 \\ \bottomrule
\end{tabular}
\caption{Official results on the AraGenEval test set. XLM-RoBERTa achieved the best performance across all metrics.}
\label{tab:main_results}
\end{table*}

\section{Experimental Setup}
\subsection{Data Splits}
The experimental setups for data splitting differed:
\begin{itemize}
    \item \textbf{AraELECTRA \& CAMeLBERT}: We used the entire training dataset of 4,734 samples for both training and evaluation during the development phase.
    \item \textbf{XLM-RoBERTa}: We split the main training data into an 80\% training set (3,787 samples) and a 20\% validation set (947 samples), stratified to maintain the label distribution.
\end{itemize}
All models were then used to generate predictions for the official `test\_unlabeled.csv` file.

\subsection{Hyperparameters}
Models were fine-tuned using the Hugging Face `transformers` library \cite{wolf-etal-2020-transformers}. Key hyperparameters are detailed in Table \ref{tab:hyperparams}.

\begin{table}[htbp]
\centering
\small
\begin{tabular}{@{}ll@{}}
\toprule
\textbf{Hyperparameter} & \textbf{Value} \\ \midrule
Learning Rate           & 2e-5 \\
Batch Size (per device) & 4 \\
Optimizer               & AdamW \\
Weight Decay            & 0.01 \\
Max Sequence Length     & 512 \\
Epochs (AraELECTRA)     & 4 \\
Epochs (CAMeLBERT)      & 4 \\
Epochs (XLM-RoBERTa)    & 5 \\ \bottomrule
\end{tabular}
\caption{Key hyperparameters for fine-tuning.}
\label{tab:hyperparams}
\end{table}

\subsection{Evaluation Metrics}
The primary metric was the macro F1-score. We also report accuracy, precision, recall, specificity, and balanced accuracy as provided by the official evaluation script.

\section{Results}
\subsection{Quantitative Findings}
Our systems yielded varied performance on the official test set, with XLM-RoBERTa emerging as the strongest model. The final results are summarized in Table \ref{tab:main_results}, which led to our 5th place finish.

\subsection{Analysis}
The most significant finding is that the multilingual XLM-RoBERTa model outperformed both specialized Arabic models. This suggests that the broader and more diverse pretraining corpus of XLM-R may have equipped it with more generalizable features for distinguishing the subtle artifacts of machine generation. As our data analysis showed, the human and machine classes have distinct lexical profiles; XLM-R's exposure to a vast range of topics and styles in 100 languages likely made it more adept at capturing these stylistic and topical differences.

In contrast, AraELECTRA performance was notably lower. We hypothesize that our aggressive text normalization and diacritic removal, intended to simplify the task, was detrimental. By stripping these features, we likely removed fine-grained signals (e.g., stylistic choices in vocabulary, specific named entities) that our data analysis identified as crucial differentiators between the news-focused human texts and the more formal machine texts. CAMeLBERT provided a strong baseline but could not match the generalization of XLM-R.

\subsection{Error Analysis}
While a detailed error analysis was not conducted, the performance gap suggests clear avenues for investigation. The lower precision of all models compared to their recall indicates a tendency to misclassify human text as machine-generated. We hypothesize that errors may stem from domain mismatch or from human-written text that is formulaic or stylistically simple, thus resembling patterns typical of AI generation. Future work should focus on a qualitative analysis of these false positives.

\section{Conclusion}
In this paper, we presented our comparative approach for the AraGenEval Shared Task, which resulted in a 5th place ranking. Our experiments showed that the multilingual XLM-RoBERTa model is surprisingly effective for Arabic AI-generated text detection, outperforming specialized monolingual models. Our data analysis revealed significant differences in text length and lexical choice between classes, which likely played a key role in model performance.

Our primary limitation was the suboptimal performance of the AraELECTRA model, likely due to a counterproductive preprocessing strategy. Future work should explore less aggressive text normalization, experiment with model ensembling, and perform a detailed error analysis to better understand the failure modes on this nuanced task.

\section*{Acknowledgments}
This research is supported by the Higher Education Commission (HEC), Government of Pakistan, under the National Research Program for Universities (NRPU), titled “Automatic Multi-Model Classification of Religious Hate Content from Social Media.” The work is conducted at the National University of Computer and Emerging Sciences, Karachi Campus, under Grant NRUP-16153.
We would also like to thank the organizers of the AraGenEval Shared Task for providing the dataset and the opportunity to participate.

\bibliography{custom}

\begin{thebibliography}{10}
\providecommand{\natexlab}[1]{#1}

\bibitem[{Abudalfa et~al.(2025)Abudalfa, Ezzini, Abdelali, Alami, Benlahbib,
  Chafik, El-Haj, El~Mahdaouy, Jarrar, Lamsiyah, and
  Luqman}]{abudalfa2025arageneval}
Shadi Abudalfa, Saad Ezzini, Ahmed Abdelali, Hamza Alami, Abdessamad Benlahbib,
  Salmane Chafik, Mo~El-Haj, Abdelkader El~Mahdaouy, Mustafa Jarrar, Salima
  Lamsiyah, and Hamzah Luqman. 2025.
\newblock The arageneval shared task on arabic authorship style transfer and
  ai-generated text detection.
\newblock In \emph{Proceedings of the Third Arabic Natural Language Processing
  Conference (ArabicNLP 2025)}, Suzhou, China. Association for Computational
  Linguistics.

\bibitem[{Al-Shboul et~al.(2024)Al-Shboul, Al-Tarawneh, Al-Shboul, and
  Al-Shboul}]{alshboul2024comprehensive}
Ibrahim Al-Shboul, Moath Al-Tarawneh, Ahmad Al-Shboul, and Anas Al-Shboul.
  2024.
\newblock \href {https://doi.org/10.3390/eng8030032} {A comprehensive overview
  of arabic text classification using deep learning models}.
\newblock \emph{Eng}, 8(3):32.

\bibitem[{Antoun et~al.(2021)Antoun, Baly, and Hajj}]{antoun2021araelectra}
Wissam Antoun, Fady Baly, and Hazem Hajj. 2021.
\newblock Araelectra: Pre-training text discriminators for arabic language
  understanding.
\newblock In \emph{Proceedings of the sixth Arabic natural language processing
  workshop}, pages 191--201.

\bibitem[{Clark et~al.(2020)Clark, Luong, Le, and Manning}]{clark2020electra}
Kevin Clark, Minh-Thang Luong, Quoc~V Le, and Christopher~D Manning. 2020.
\newblock Electra: Pre-training text encoders as discriminators rather than
  generators.
\newblock In \emph{International Conference on Learning Representations}.

\bibitem[{Conneau et~al.(2020)Conneau, Khandelwal, Goyal, Chaudhary, Wenzek,
  Guzm{\'a}n, Grave, Ott, Zettlemoyer, and Stoyanov}]{conneau2019unsupervised}
Alexis Conneau, Kartikay Khandelwal, Naman Goyal, Vishrav Chaudhary, Guillaume
  Wenzek, Francisco Guzm{\'a}n, Edouard Grave, Myle Ott, Luke Zettlemoyer, and
  Veselin Stoyanov. 2020.
\newblock Unsupervised cross-lingual representation learning at scale.
\newblock In \emph{Proceedings of the 58th Annual Meeting of the Association
  for Computational Linguistics}, pages 8440--8451.

\bibitem[{Devlin et~al.(2019)Devlin, Chang, Lee, and
  Toutanova}]{devlin2018bert}
Jacob Devlin, Ming-Wei Chang, Kenton Lee, and Kristina Toutanova. 2019.
\newblock Bert: Pre-training of deep bidirectional transformers for language
  understanding.
\newblock In \emph{Proceedings of the 2019 Conference of the North American
  Chapter of the Association for Computational Linguistics: Human Language
  Technologies, Volume 1 (Long and Short Papers)}, pages 4171--4186.

\bibitem[{Inoue et~al.(2021)Inoue, Al-Rifou, and Habash}]{inoue2021interplay}
Go~Inoue, Bashar Al-Rifou, and Nizar Habash. 2021.
\newblock The interplay of variant, genre, and domain for arabic text
  classification.
\newblock In \emph{Proceedings of the sixth Arabic natural language processing
  workshop}, pages 1--15.

\bibitem[{Vaswani et~al.(2017)Vaswani, Shazeer, Parmar, Uszkoreit, Jones,
  Gomez, Kaiser, and Polosukhin}]{vaswani2017attention}
Ashish Vaswani, Noam Shazeer, Niki Parmar, Jakob Uszkoreit, Llion Jones,
  Aidan~N Gomez, {\L}ukasz Kaiser, and Illia Polosukhin. 2017.
\newblock Attention is all you need.
\newblock In \emph{Advances in neural information processing systems}, pages
  5998--6008.

\bibitem[{Wolf et~al.(2020)Wolf, Debut, Sanh, Chaumond, Delangue, Moi, Cistac,
  Rault, Louf, Funtowicz, Davison, Shleifer, von Platen, Ma, Jernite, Plu, Xu,
  Le~Scao, Gugger, Drame, Lhoest, and Rush}]{wolf-etal-2020-transformers}
Thomas Wolf, Lysandre Debut, Victor Sanh, Julien Chaumond, Clement Delangue,
  Anthony Moi, Pierric Cistac, Tim Rault, R{\'e}mi Louf, Morgan Funtowicz, Joe
  Davison, Sam Shleifer, Patrick von Platen, Clara Ma, Yacine Jernite, Julien
  Plu, Canwen Xu, Teven Le~Scao, Sylvain Gugger, and 3 others. 2020.
\newblock \href {https://www.aclweb.org/anthology/2020.emnlp-demos.6}
  {Transformers: State-of-the-art natural language processing}.
\newblock In \emph{Proceedings of the 2020 Conference on Empirical Methods in
  Natural Language Processing: System Demonstrations}, pages 38--45, Online.
  Association for Computational Linguistics.

\bibitem[{Zain et~al.(2025)Zain, Farooqui, and Rafi}]{busted2025multi}
Ali Zain, Sareem Farooqui, and Muhammad Rafi. 2025.
\newblock A multi-strategy approach for ai-generated text detection.
\newblock In \emph{Proceedings of the 15th International Conference on Recent
  Advances in Natural Language Processing (RANLP)}.

\end{thebibliography}

\end{document}